\documentclass[12pt]{colt2019} 

\title[Graph Laplacian Regularizer]{Error Analysis on Graph Laplacian Regularized Estimator}
\usepackage{times, xcolor}
\usepackage{enumitem}

\coltauthor{%
 \Name{Kaige Yang} \Email{kaige.yang.11@ucl.ac.uk}\\
 \addr University College London
 \AND 
 \Name{Xiaowen Dong} \Email{xiaowen.dong@eng.ox.ac.uk}\\
 \addr Oxford Univeristy%
\AND
 \Name{Laura Toni}
 \Email{l.toni@ucl.ac.uk}\\
 \addr University College London\\
 }

\begin{document}

\maketitle

\begin{abstract}
We provide a theoretical analysis of the representation learning problem aimed at learning the latent variables (design matrix) $\Theta$ of observations $Y$ with the knowledge of the coefficient matrix $X$. The design matrix is learned under the assumption that the latent variables $\Theta$ are smooth with respect to a (known) topological structure $\mathcal{G}$. To learn such latent variables, we study a graph Laplacian regularized estimator, which is the penalized least squares estimator with penalty term proportional to a Laplacian quadratic form. This type of estimators has recently received considerable attention due to its capability in incorporating underlying topological graph structure of variables into the learning process. While the estimation problem can be solved efficiently by state-of-the-art optimization techniques, its statistical consistency properties have been largely overlooked. In this work, we develop a non-asymptotic bound of estimation error under the classical statistical setting, where sample size is larger than the ambient dimension of the latent variables. This bound illustrates  theoretically the impact of the alignment between the data and the graph structure as well as the graph spectrum on the estimation accuracy. It also provides theoretical evidence of the advantage, in terms of convergence rate, of the graph Laplacian regularized estimator over classical ones (that ignore the graph structure) in case of a smoothness prior. Finally, we provide empirical results of the estimation error to corroborate the theoretical analysis.
\end{abstract}

\section{Introduction}
We consider the problem of representation learning with a general noisy setting, and under the classical statistical setting where the sample size $n$ is larger than the ambient dimension $p$ of 
variables. The aim of representation learning is usually to estimate the latent variables (i.e., the design matrix) $\Theta$ and the coefficient matrix $X$ that explain the intrinsic characteristics of the observations. This is usually solved by iterating between estimation of the design matrix and estimation of the coefficient matrix. In this work, we focus on the estimation of the design matrix $\Theta$ in $Y=\Theta X$, where the coefficient matrix $X$ and the observations $Y$ are assumed known, corresponding to one estimation step in representation learning. While there exists error analyses on ridge regression based estimator for the coefficient matrix, there has been little effort devoted to such analysis on the estimator for the design matrix. To provide a better understanding of the uncertainty of estimation in representation learning, it is therefore essential to derive a theoretical error analysis on the estimator for the design matrix, which is the goal of this paper.

Estimating the design matrix given the observation and the coefficient matrix is a learning problem that appears naturally and frequently in applications across many fields such as image denoising, compress sensing, dictionary learning, and collaborative filtering (CF). 
In CF \citep{rao2015collaborative}, the goal is to estimate users' preferences $\Theta$ through item features $X$ based on users' responses $Y$ (e.g., ratings). Dictionary learning \citep{yankelevsky2016dual} also shares a similar formulation, where $\Theta$ and $X$ are referred to dictionary atoms and coding coefficient matrix. In this case, our model corresponds to the case when the coding coefficient matrix $X$ is known. Another closely related application is image denoising \citep{wang2018marginalized}, where $\Theta$ corresponds to a collection of basis functions (\textit{e.g.}, wavelets, cosine waves) modeling the image signal as a linear combination of these basis functions with $X$ being the coefficients. 

Due to the wide applicability of representation learning problems, the development of estimators for representation learning has recently received a substantial attention. A key type of estimators is regularized least squares estimators, which poses structural constraints
on the unknown coefficient matrix $X$.
For example, the Lasso formulation \citep{tibshirani1996regression} poses a sparsity constraint on the number of non-zero entry in $\Theta$. 
The work of \citet{bach2008consistency} introduces nuclear/trace norm based regularizers aiming to find a low rank solution while fitting the data. Other examples include \citet{zhou2014regularized} and \citet{ huang2011learning} that are based on various structural constraints. 
Compared to estimating the coefficient matrix, however, much less work has been devoted to the estimation of the design matrix, and in particular error analysis associated with the uncertainty in such estimation.

In this paper, we study the estimation of the design matrix in representation learning, where the estimator involves a graph based regularizer. Such a regularizer helps incorporating the underlying geometric structure of the data into the learning process.
For example, in recommender systems \citep{eirinaki2018recommender}, one might get access to users' social network. Incorporating such information may lead to a better understanding of users' preference ($\Theta$), which may in turn improve the recommendation performance ($Y$).
We focus on the graph Laplacian based regularizer, which has been widely adopted in the literature \citep{huang2018matrix,rao2015collaborative,dadkhahi2018alternating,yankelevsky2016dual} thanks to its mathematical regularity (\textit{e.g.,} convexity and differentiability). In particular, the least squares estimation regularized by $tr(\Theta^T L \Theta)$ is a convex program, and $\Theta$ can be computed by the well known Bartels-Stewart algorithm \citep{bartels1972solution} or more efficient algorithms developed recently \citep{rao2015collaborative,ji2018learning}.
Rather than focusing on solving  methods for the estimation problem, our goal is rather to study its statistical consistency guarantee, and to gain insights about the impact of $L$ on its convergence rate. We aim to provide a bound on the estimation error, $\bigtriangleup=\hat{\Theta}-\Theta^*$, defined as the difference between any estimation $\hat{\Theta}$ and the unknown groundtruth latent variables $\Theta^*$. More formally, we derive a non-asymptotic upper bound on $||\bigtriangleup||_F=||\hat{\Theta}-\Theta^*||_F$ that holds with high probability, where $||\cdot||_F$ is the Frobenius norm. To the best of our knowledge, this theoretical analysis is absent in the literature.  

The main contributions of the paper are as follows:
\begin{itemize}[noitemsep]
    \item we obtain an error bound for the graph Laplacian regularized estimator, which illustrates the effect of graph structure as well as its alignment with data on estimation accuracy; 
    \item we show the impact of the graph spectrum (\textit{i.e.,} eigenvalues of the graph Laplacian) on the estimation accuracy;
    \item we compare with the classical ridge estimator to prove theoretically advantages brought by incorporating graph structure into the estimator; 
    \item we validate our claims empirically by simulations.
\end{itemize}
In summary, we study analytically the estimation of the design matrix to understand the associated uncertainty. This is a key component to understand the effectiveness of representation learning algorithms, and to understand the effect of the topological structure $\mathcal{G}$ on the estimation uncertainty.

The reminder of the paper is organized as follows. Section 2 presents related work and in particular existing studies on the theoretical analysis of representation learning. Section 3 introduces the basic definition related to the graph Laplacian regularized estimator, and formulates the estimation problem. Section 4 and 5 present the main results of error analysis on the estimator and the corresponding proofs, respectively. Section 6 shows empirical results and Section 7 summarizes the paper.

\

\textbf{Notations:} Let $A=[A_{ij}] \in \mathbb{R}^{n \times k}$ and $B=[B_{ij}]\in \mathbb{R}^{n \times k}$ be two $n \times m$ matrices. The scalar product $\langle A, B\rangle =tr(A^TB)$, where $tr(\cdot)$ is the trace operator. The Frobenius norm is defined as  $||A||_F=(\sum_{ij}A_{ij}^2)^{\frac{1}{2}}$ and the infinity norm as $||A||_\infty=\max_{ij}|A_{ij}|$. The nuclear norm is defined as $||A||_*=tr(\sqrt{A^TA})$, with $A^T$ being the transpose of $A$. Let $\mathbf{x}=[x_1,..., x_d]^T \in \mathbb{R}^d$ be a \textit{d}-dimensional vector, where its $L_1$ and $L_2$ norms are defined as $||\mathbf{x}||_1=\sum_{i=1}^d|x_i|$ and $||\mathbf{x}||_2=(\sum_{i=1}^d x_i^2)^{\frac{1}{2}}$, respectively.

\section{Related work}
In many applications, data come with an underlying geometric structure, typically in the form of a graph, which should be taken into account in the learning process. 
There has been an increasing amount of interest in representation learning, where topological graph structures are embedded into estimators to promote desirable properties of the solution. For example, \citet{zhou2004regularization} introduces a measure of smoothness of the data with respect to a graph topology, in the form of the so-called Laplacian quadratic form $tr(\Theta^TL\Theta)$. Employing this term as a regularizer in the estimators thus finds a solution $\hat{\Theta}$ that is smooth on the graph. Alternatively, \citet{shuman2012signal} introduces total variation and graph total variation estimation. Following works show empirically the effectiveness of such regularizers \citep{kalofolias2014matrix,rao2015collaborative,zhao2015expert}. Other graph-based regularizers include edge Lasso \citep{sharpnack2012sparsistency}, network Lasso \citep{hallac2015network}, and graph trend filtering \citep{wang2016trend}.

In this work, we study a graph Laplacian regularized least squares estimator for the design matrix $\Theta$. To the best of our knowledge, there is no prior work on the theoretical understanding of a consistency guarantee of this estimator, either in high-dimensional ($n<p$, or $n \ll p)$ or in classical statistical setting ($n \geq p)$.
Nevertheless, there has been a few theoretical studies on the graph regularized estimators. For example, total variation regularized estimators \citep{shuman2012signal} are proven to be similar to graph Laplacian regularized estimators \citep{kalofolias2014matrix} and promote piece-wise constant solutions, while the Laplacian regularized estimators lead to piece-wise smooth solutions.
The work of \citet{hutter2016optimal} provides optimal rate analysis of the total variation regularized estimator. The work of \citet{rao2015collaborative} derives statistical consistency guarantees of the Laplacian regularized estimator in the application of collaborative filtering.
A key difference from their paper is that,
in our work, we consider the unknown coefficient matrix $\Theta$ as the graph signals.
Another difference is that, they derive a bound on the prediction error in the measurements $||\hat{Y}-Y^*||_F$, while we develop a bound on the estimation error in the design matrix $||\hat{\Theta}-\Theta^*||_F$. Finally, the work of \citet{li2018graph} provides theoretical consistency guarantees on a graph regularized estimator in linear regression, which is formulated using a combination of graph Laplacian, total variation, and edge Lasso. Instead, we focus on such theoretical properties using only graph Laplacian based regularizer.

\section{Graph Laplacian regularized estimator}
In this section, we introduce our estimation problem in representation learning. We first provide some background on signals on graphs, and then introduce 
the graph Laplacian regularized estimator and the associated estimation problem.

\subsection{Graph Laplacian and graph signal}
Consider a weighted and undirected graph $\mathcal{G}=(V, E, W)$ of $m$ vertices, where $V$ is the finite set of vertices and $E$ the finite set of edges, and $W=[W_{ij}] \in \mathbb{R}^{m \times m}$ denotes the weighted adjacency matrix. The entry $W_{ij}$ represents the edge weight between vertex $v_i$ and $v_j$. $W_{ij}=0$ if $v_i$, $v_j$ are not directly connected, and $W_{ij} > 0$ if connected. Moreover, $W_{ij}=W_{ji}$ for weighted undirected graph. The graph degree matrix is $D=[D_{ii}] \in \mathbb{R}^{m \times m}$, where $D_{ii}=\sum_jW_{ij}$ represents the degree of vertex $v_i$. The combinatorial graph Laplacian $L$ is defined to be $L=D-W$.

A graph signal is referred as a function $f:V \rightarrow \mathbb{R}^m$ that assigns a real value to each graph vertex. In this paper, we consider smooth signals over graphs. With the Laplacian matrix $L$, the smoothness of signal $f$ over graph $\mathcal{G}$ can be quantified as a quadratic form of $L$ \citep{zhou2004regularization}:
\begin{equation}
    f^TLf=\frac{1}{2}\sum_{i \sim j}W_{ij}(f(i)-f(j))^2
\end{equation}
which is a weighted sum of the squared signal difference between connected vertices, where weights are corresponding edge weights.

\subsection{The estimation problem}
We now state the estimation problem to be addressed in this paper, which is the estimation of the design martix $\Theta$, in a linear model with a graph Laplacian regularizer. We first introduce the linear model, and then describe the graph Laplacian regularized estimator.

We consider the problem of representation learning under a general noisy setting. The aim of this problem is to learn the design matrix $\Theta \in \mathbb{R}^{m \times k}$ that explains the observations $Y \in \mathbb{R}^{m \times n}$ with the coefficient matrix $X \in \mathbb{R}^{k \times n}$. Formally, we consider a linear model in the form of 
\begin{equation}
    Y=\Theta X+\Omega
\end{equation}
where $Y\in \mathbb{R}^{m \times n}$ denotes the observation matrix, $X \in \mathbb{R}^{k \times n}$ is the coefficient matrix  following the standard regularity assumption that columns are independent, and $\Theta \in \mathbb{R}^{m \times k}$ represents the design matrix. Let us denote the noise matrix by $\Omega=[\Omega_{ij}] \in \mathbb{R}^{m \times n}$ with entries $\Omega_{ij} \sim \mathcal{N}(0, \sigma^2)$ being Gaussian noise.
We now consider an estimator arising frequently in the literature, which assumes $\Theta$ to be smooth with respect to an underlying graph structure $\mathcal{G}$.

To estimate   $\Theta$ under the prior of smoothness on $\mathcal{G}$, we consider a graph Laplacian regularized estimator. Formally, 
\begin{equation}
    \hat{\Theta}=\arg \min_{\Theta \in \mathbb{R}^{m \times k}} \frac{1}{2n}||Y-\Theta X||_F^2+\alpha\  tr(\Theta^TL\Theta)
\end{equation}
where $\alpha \geq 0$ is a regularization parameter. The regularization term $tr(\Theta^TL\Theta)$ is known to promote smoothness of $\Theta$ with respect to the underlying graph $\mathcal{G}$. From the perspective of statistical models, the regularizer in (3) assumes that 
$\Theta$ follows a degenerate multivariate Gaussian distribution, where the graph Laplacian $L$ acts as the precision matrix. Such an estimator has been adopted in applications such as graph-structured matrix factorization \citep{kalofolias2014matrix, huang2018matrix} and signal denoising \citep{pang2017graph}.

It is instructive to compare (3) with a simpler estimator which applies the standard ridge estimator to the representation learning problem (2), solving
\begin{equation}
    \hat{\Theta}=\arg\min_{\Theta \in \mathbb{R}^{m \times k}} \frac{1}{2n}||Y-\Theta X||_F^2+\alpha \ tr(\Theta^T I_m \Theta)
\end{equation}\\
where $I_m$ is a $m \times m$ identity matrix. It is worth noting that the estimator (4) is a degenerative version of (3) when $L=I_m$, \textit{i.e.,} when the graph structure is ignored. Therefore, the estimator in (4) is a desirable baseline for understanding the property of the graph Laplacian regularized estimator in (3). In the following sections, we provide a deep analysis on (3) as well as a comparison with (4).

\section{Estimation error analysis}
The central focus of the paper is the error analysis on $\hat{\Theta}$, i.e., providing a bound on the estimation error $\bigtriangleup=\hat{\Theta}-\Theta^*$, where $\Theta^*$ is the unknown ground-truth latent variables matrix.
To derive this bound, we generally follow the unified analysis framework of \citet{negahban2012unified}, properly adjusted to our problem. We first develop some key notations of the graph Laplacian regularized estimator in (3).
Next, we describe an important ingredient of our main result: the \textit{strong convexity} condition. Finally, we present our main results and their interpretation.

\subsection{Key notations}
The graph Laplacian regularized estimator (3) can be rewritten into the following form:
\begin{equation}
    \hat{\Theta}=\arg\min_{\Theta \in \mathbb{R}^{m \times k}} \mathcal{L}(\Theta)+\alpha \mathcal{R}(\Theta)
\end{equation}
where $\mathcal{L}(\Theta)=\frac{1}{2n}||Y-\Theta X|_F^2$ is the loss function assigning a cost to any $\Theta \in \mathbb{R}^{m \times k}$ given a pair of $\{X, Y\}$, and $\mathcal{R}(\Theta)=tr(\Theta^TL\Theta)$ denotes the Laplacian regularizer.\\
The first order Taylor expansion of the loss function at $\Theta^*$ in the direction $\bigtriangleup=\hat{\Theta}-\Theta^*$ is expressed as 
\begin{equation}
    \mathcal{L}(\Theta^*+\bigtriangleup)=\mathcal{L}(\Theta^*)+\langle \bigtriangledown \mathcal{L}(\Theta^*), \bigtriangleup \rangle
\end{equation}
thus, the error of the first Taylor expansion $\delta \mathcal{L}(\Theta^*)$ is defined as 
\begin{equation}
    \delta \mathcal{L}(\Theta^*)=\mathcal{L}(\Theta^*+\bigtriangleup)-\mathcal{L}(\Theta^*)-\langle \bigtriangledown \mathcal{L} (\Theta^*), \bigtriangleup \rangle
\end{equation}
After some algebraic steps (proved in Appendix A), we can also observe that
\begin{equation}
    \bigtriangledown\mathcal{L}(\Theta^*)=\frac{1}{n}\Omega X^T
\end{equation}
\begin{equation}
    \delta\mathcal{L}(\Theta^*)=\frac{1}{2n}||\bigtriangleup X||_F^2
\end{equation}
where $\Omega=\Theta^*X-Y$.\\
\subsection{Strong convexity}
We now pose a technical condition on the error of the Taylor expansion, $\delta \mathcal{L}(\Theta^*)$, which provides a desirable control of the error magnitude. This bound is based on the  \textit{strong convexity} condition \citep{negahban2012restricted}, formally expressed as follows 
\begin{equation}
    \delta \mathcal{L}(\Theta^*)\geq \kappa ||\bigtriangleup||_F^2, \ \ \ \text{for} \bigtriangleup \text{ around } \ \Theta^*
\end{equation}
where $\kappa>0$ is a positive constant. Intuitively, (10) requires the loss function $\mathcal{L}$ is sharply curved around its optimal solution $\hat{\Theta}$ by setting a lower bound on its gradient. The necessity of this requirement can be interpreted as follows. Consider the difference loss $\mathcal{L}(\hat{\Theta})-\mathcal{L}(\Theta^*)$, it is expected that small $\mathcal{L}(\hat{\Theta})-\mathcal{L}(\Theta^*)$ indicates small $\bigtriangleup=\hat{\Theta}-\Theta^*$. However, this assumption is reasonable only when $\mathcal{L}$ sharply curves at $\hat{\Theta}$. For example, to illustrate this point, if $\mathcal{L}$ is relative flat curved, a large $\bigtriangleup=\hat{\Theta}-\Theta^*$ might also leads to small $\mathcal{L}(\hat{\Theta})-\mathcal{L}(\Theta^*)$. Therefore, to avoid the curve of $\mathcal{L}$ is too flat, we pose a strong convexity (10) constraint on it , which provides a desirable control of the magnitude of the error $\bigtriangleup$ in turn.\\
\newline
Note that in \cite{negahban2011estimation} a similar notion named \textit{restricted strong convexity} is introduced. The term ``restricted" means a constraint on the set of $\bigtriangleup$, which is necessary in high-dimensional statistical inference, 
in settings in which   the ambient dimension $k$ is larger than the sample size $n$ (\textit{i.e, $k \geq n$ or $k \gg n$}). In such settings, the global strong convexity is not always ensured, thus it is necessary to restrict $\bigtriangleup$ into a set  where the \textit{strong convexity} holds (hence the restricted strong convexity). In contrast, in this paper we consider the standard setting, $n > k$, where it is natural to assume the loss function is strongly convex at a global scale under mild conditions. 
Another analogous condition known as ``restricted eigenvalues'' (RE) is introduced in \cite{rohde2011estimation}. 

\subsection{Main results}
Equipped with the above notations and assumptions, we are ready to state our main result: a deterministic upper bound on the estimation error of the Laplacian regularized estimator (3), which holds with high probability and it is defined in the following theorem.
\begin{theorem}
Consider the linear model (2), where the strong convexity condition (10) holds and the regularization parameter $\alpha \geq 8\sigma \sqrt{D}\frac{\sqrt{m+k}}{mn}$ with any constant $D\geq 2$. Imposing $rank(\bigtriangleup) \leq r$, then the optimal solution $\hat{\Theta}$ obtained by (3) satisfies the following error bound:
\begin{equation}
    ||\hat{\Theta}-\Theta^*||_F \leq \frac{\alpha(\sqrt{r}+2||L\Theta^*||_F)}{\kappa+\alpha \lambda_2}
\end{equation}
with high probability. where $\lambda_2$ is the second smallest eigenvalue of the graph Laplacian $L$.
\end{theorem}
The sketched proof of \textbf{Theorem 1} is presented in Section 5.2, while detailed proof is postponed to Appendix B. In the following, we provide key interpretations of \textbf{Theorem 1}.
\begin{enumerate}
    \item[$(a)$] Note that \textbf{Theorem 1} is a non-asymptotic bound on the optimas of (3) given a fixed regularization parameter $\alpha$. When applied to particular models, the \textit{strong convexity} condition and the assumption $\alpha \geq ||\bigtriangledown \mathcal{L}(\Theta^*)||_\infty$ are required to be satisfied.
    \item[$(b)$]The term $||L\Theta^*||_F$ quantifies the alignment between $\Theta^*$ and graph information $L$. Suppose there exists a groundtruth $L^*$ over which $\Theta^*$ is smooth. It can be verified that any deviation of $L$ from $L^*$ would lead to a larger value of $||L\Theta^*||_F$ than $||L^*\Theta^*||_F$. In other words,
    deviations of $\Theta^*$ from the smoothness assumption leads to a larger estimation error. 
    \item[$(c)$] The term $\lambda_2$ illustrates the impact of the density of graph on the estimation accuracy. By comparing the eigenvalue profiles of two graphs, it can be seen that $\lambda_2$ of a dense graph is typically larger than that of a sparse graph. Therefore, from (11), we can deduce that dense graph leads to lower error, if it indeed aligns with $\Theta^*$, because that a dense graph indicates more correlated rows of $\Theta^*$, with the help of $L$, the Laplacian regularized estimator is expected to result in more accurate estimation.
\end{enumerate}
Finally, being the estimator defined in (4) is shown to be a degenerative case of (3), the bound (11) applies to (4) as well. The following corollary provides a bound applied to (4), with the detailed proof provided in Appendix C.\\
\newline
\textbf{Corollary 1}
\textit{
Consider the linear model (2), where the strong convexity condition (10) holds and the regularization parameter $\alpha \geq 8\sigma \sqrt{D}\frac{\sqrt{m+k}}{mn}$ with any constant $D\geq 2$. If the rank of $\bigtriangleup=\hat{\Theta}-\Theta^*$ is at most $r$. Then the optimal solution $\hat{\Theta}$ obtained by the ridge estimator (4) satisfies the following error bound:}
\begin{equation}
    ||\hat{\Theta}-\Theta^*||_F \leq \frac{\alpha(\sqrt{r}+2||\Theta^*||_F)}{\kappa+\alpha}
\end{equation}
with high probability.\\
\textbf{Corollary 1} takes a simpler form than \textbf{Theorem 1} since $L$ is ignored and $\lambda_2(I_m)=1$.
\section{Proofs}
In this section, we sketch the proofs of \textbf{Theorem 1}, while more detailed proof is provided in Appendix B. We first present some Lemmas used in the proof.
\subsection{Assumption and Lemmas}
\textbf{Assumption 1} \textit{Given the definition $\bigtriangleup=\hat{\Theta}-\Theta^*$, we assume $\bigtriangleup$ satisfies the following property}
\begin{equation}
    \sum_{j=1}^k\sum_{i=1}^m\bigtriangleup_{ji}^2\gg \frac{1}{m}\sum_{j=1}^k(\sum_{i=1}^m\bigtriangleup_{ji})^2)
\end{equation}
See Appendix D for detailed reasoning.\\
\newline
\textbf{Lemma 1} \textit{
Let the eigenvalues of the graph Laplacian $L$ be denoted by $0=\lambda_1\leq \lambda_2 \leq... \leq \lambda_m$. Suppose $\bigtriangleup$ satisfies \textbf{Assumption 1}, then $tr(\bigtriangleup^TL\bigtriangleup)$ satisfies the following lower bound}
\begin{equation}
    tr(\bigtriangleup^TL\bigtriangleup)\geq \lambda_2||\bigtriangleup||_F^2
\end{equation}
The detailed proof is provided in Section 5.3.\\
\newline
\textbf{Lemma 2} \citep{rohde2011estimation}.\textit{
Let entries of $\Omega=[\Omega_{ij}]$ be \textit{i.i.d.} $\mathcal{N}(0, \sigma^2)$ random variables, where $\Omega \in \mathbb{R}^{m \times n}$. $X\in \mathbb{R}^{k \times n}$ follows standard statistical regularity with independent columns. Then, for any $D\geq 2$,}
\begin{equation}
    \frac{1}{n}||\Omega X^T||_\infty \leq 8\sigma \sqrt{D}\frac{\sqrt{m+k}}{mn}
\end{equation}
with probability at least $1-2exp(-(D-log5)(m+k)$.\\
\newline 
\textbf{Lemma 3} \citep{negahban2011estimation}. \textit{
Let $X \in \mathbb{R}^{k \times n}$ be a random matrix with \textit{i.i.d.} columns sampled from a \textit{k}-variate $\mathcal{N}(0, \Sigma)$. Then for $n \geq k$, we have}
\begin{equation}
    \mathbb{P}[\sigma_{min}(\frac{1}{n}XX^T)\geq \frac{\sigma_{min}(\Sigma)}{9}, \sigma_{max}(\frac{1}{n}(XX^T) \leq 9 \sigma_{max}(\Sigma)] \geq 1-4exp(-n/2)
\end{equation}
\subsection{Sketch proof of Theorem 1}
Due to the optimality of $\hat{\Theta}$ for (5), we have 
\begin{equation}
\mathcal{L}(\hat{\Theta})+\alpha \mathcal{R}(\hat{\Theta})\leq \mathcal{L}(\Theta^*)+\alpha \mathcal{R}(\Theta^*)
\end{equation}
Substituting $\hat{\Theta}=\Theta^*+\bigtriangleup$ and $\mathcal{R}(\Theta)=tr(\Theta^TL\Theta)$ yields
\begin{equation}
    \mathcal{L}(\Theta^*+\bigtriangleup)-\mathcal{L}(\Theta^*)+\alpha(2 tr((\Theta^*)^TL\bigtriangleup)+tr(\bigtriangleup^TL\bigtriangleup)) \leq 0
\end{equation}
By the definition of scalar product and its property, we have 
\begin{equation}
    tr((\Theta^*)^TL\bigtriangleup)=\langle L\Theta^*, \bigtriangleup\rangle \geq -|\langle L\Theta^*, \bigtriangleup\rangle|
\end{equation}
Combining this with (7) the definition of $\delta\mathcal{L}(\Theta^*)$ and $\langle \bigtriangledown \mathcal{L}(\Theta^*), \bigtriangleup \rangle \geq -|\langle \bigtriangledown \mathcal{L}(\Theta^*), \bigtriangleup \rangle|$, we know that 
\begin{equation}
-|\langle \bigtriangledown \mathcal{L}(\Theta^*), \bigtriangleup \rangle|+\delta \mathcal{L}(\Theta^*)+\alpha (-2|\langle L\Theta^*, \bigtriangleup\rangle|+tr(\bigtriangleup^TL\bigtriangleup))\leq 0
\end{equation}
Applying the H\"{o}lder Inequality \citep{kuptsov2001holder}, we have $|\langle \bigtriangledown \mathcal{L}( \Theta^*), \bigtriangleup \rangle|\leq ||\bigtriangledown \mathcal{L}( \Theta^*)||_{\infty} ||\bigtriangleup||_*$ and $|\langle L \Theta^*, \bigtriangleup \rangle|\leq ||L\Theta^*||_F ||\bigtriangleup||_F$. Thus
\begin{equation}
    \delta \mathcal{L}(\Theta^*, \bigtriangleup)+\alpha tr(\bigtriangleup^TL\bigtriangleup)\leq ||\bigtriangledown\mathcal{L}(\bigtriangleup, \Theta^*)||_\infty||\bigtriangleup||_*+2\alpha ||L\Theta^*||_F ||\bigtriangleup||_F
\end{equation}
Imposing the \textit{strong convexity} condition $\delta \mathcal{L}(\Theta^*)\geq \kappa||\bigtriangleup||_F^2$. Assume $\alpha \geq ||\bigtriangledown \mathcal{L}(\Theta^*)||_{\infty}$. Note the fact that if $rank{(\bigtriangleup)}\leq r$, then $ ||\bigtriangleup|_*\leq \sqrt{r} ||\bigtriangleup||_F$, we have 
\begin{equation}
    \kappa||\bigtriangleup||_F^2+\alpha tr(\bigtriangleup^TL\bigtriangleup)\leq \alpha\sqrt{r}||\bigtriangleup||_F+2\alpha||L\Theta^*||_F||\bigtriangleup||_F
\end{equation}
The remaining is to lower bound $tr(\bigtriangleup^TL\bigtriangleup)$. \textbf{Lemma 1} provides a proper lower bound on this.\\
Substituting (14) into (22) and dividing both sides with $||\bigtriangleup||_F$ yields
\begin{equation}
    ||\bigtriangleup||_F \leq \frac{\alpha(\sqrt{r}+2||L\Theta^*||_F)}{\kappa+\alpha \lambda_2}
\end{equation}
\newline
The remained issue is to choose valid values for the regularization parameter $\alpha$ and the positive constant $\kappa$ such that bound (23) holds in high probability. For the value of $\alpha$, we follow \textbf{Lemma 2}, from \cite{rohde2011estimation}, which provides a upper bound on $||\bigtriangledown \mathcal{L}(\Theta^*)||_\infty$. For the choice of $\kappa$, \textbf{Lemma 3}, obtained from \cite{negahban2011estimation}, provides a lower bound on $\delta \mathcal{L}(\Theta^*)$.
Interested readers are referred to proofs in their original work.\\
\newline 
More specifically, to decide a proper choice of $\alpha$, we need to upper bound $||\bigtriangledown \mathcal{L}(\theta^*)||_\infty$ since we assume $\alpha \geq ||\bigtriangledown \mathcal{L}(\theta^*)||_\infty$. Recall (8), we have $||\bigtriangledown \mathcal{L}(\Theta^*)||_\infty=\frac{1}{n}||\Omega X^T||_\infty$.  From \textbf{Lemma 2},
it can be seen that the choice $\alpha \geq 8\sigma \sqrt{D}\frac{\sqrt{m+k}}{mn}$ is suffice to ensure $\alpha \geq ||\bigtriangledown\mathcal{L}(\Theta^*)||_\infty$ holds in high probability.\\
\newline 
To establish the \textit{strong convexity} condition defined in (10), it is required to build a lower bound on $\delta \mathcal{L}(\Theta^*)=\frac{1}{2n}||\bigtriangleup X||_F^2$. As can be seen, similar to \cite{negahban2011estimation}, 
\begin{equation}
    \frac{1}{2n}||\bigtriangleup X||_F^2\geq \frac{\sigma_{min}(XX^T)}{2n}||\bigtriangleup||_F^2
\end{equation}
where $\sigma_{min}$ refers to the minimum singular value of the matrix $XX^T$. \textbf{Lemma 3} introduces a lower bound on $\frac{\sigma_{min}(XX^T)}{n}$. From \textbf{Lemma 3}, we can see $\frac{\sigma_{min}(XX^T)}{2n} \geq \frac{\sigma_{min}(\Sigma)}{18}$ with probability $1-4exp(-n)$. Therefore, $\kappa=\frac{\sigma_{min}(\Sigma)}{18}$ could guarantee that the condition $\delta \mathcal{L}(\Theta^*)\geq \kappa ||\bigtriangleup||_F^2$ holds with high probability.\\
With the above valid choice of $\alpha$ and $\kappa$, \textbf{Theorem 1} holds with high probability.

\subsection{Sketch proof of Lemma 1}
Let the eigendecomposition of the graph Laplacian $L$ is $L=Q\Lambda Q^T$. Define $u=Q^T\bigtriangleup$, where $u_j=Q^T\bigtriangleup_j$ and $\bigtriangleup_j$ denotes the \textit{j}-th column of $u$ and $\bigtriangleup$, respectively. 
It is straightforward to show that 
\begin{equation}
    tr(\bigtriangleup^TL\bigtriangleup)=\sum_{j=1}^k\bigtriangleup_j^TL\bigtriangleup_j=\sum_{j=1}^k\sum_{i=1}^m\lambda_iu_{ji}^2
\end{equation}
Where $\lambda_i$ denotes the \textit{i}-th eigenvalue of $L$, and $u_{ji}$ denotes the \textit{i}-th entry of the \textit{j}-th column of $u$.\\
Given $L$ is a symmetric positive semidefinite matrix, its eigenvalues are real and nonnegative. Moreover, we assume that the graph $\mathcal{G}$ is a connected component, hence $\lambda_1=0$. If we denote its eigenvalues as $0=\lambda_1 \leq \lambda_2 \leq ... \leq \lambda_m$, we have 
\begin{equation}
    \sum_{i=1}^m \lambda_iu_{ji}^2
    =\sum_{i=2}^m \lambda_iu_{ji}^2
    \geq \sum_{i=2}^m \lambda_2u_{ji}^2
    =(\sum_{i=1}^m \lambda_2u_{ji}^2)-\lambda_2u_{j1}^2
    =\lambda_2||\bigtriangleup_j||_2^2-\lambda_2u_{j1}^2
\end{equation}
 The first inequality is due to $\lambda_1=0$. Substituting (26) into (25) yields
\begin{equation}
    tr(\bigtriangleup^TL\bigtriangleup)\geq \sum_{j=1}^k(\lambda_2||\bigtriangleup_j||^2_2-\lambda_2u_{j1}^2)=\lambda_2||\bigtriangleup||_F^2-\lambda_2||Q^T_1\bigtriangleup||_2^2
\end{equation}
Where $Q^T_1=[1/\sqrt{m}, 1/\sqrt{m}, ..., 1/\sqrt{m}]^T$ is the first eigenvector of $L$.  Therefore,  $||Q^T_1\bigtriangleup||_2^2=\sum_{j=1}^k \frac{1}{m}(\sum_{i=1}^m\bigtriangleup_{ji})^2$. Also note that $||\bigtriangleup||_F^2=\sum_{j=1}^k\sum_{i=1}^m\bigtriangleup_{ji}^2$, So (27) turns to 
\begin{equation}
\begin{split}
   tr(\bigtriangleup^TL\bigtriangleup)
   &\geq \lambda_2||\bigtriangleup||_F^2-\lambda_2||Q^T_1\bigtriangleup||_2^2\\
    &=\lambda_2\sum_{j=1}^k\sum_{i=1}^m\bigtriangleup_{ji}^2-\lambda_2 \sum_{j=1}^k\frac{1}{m}(\sum_{i=1}^m\bigtriangleup_{ji})^2
\end{split}
\end{equation}
According to \textbf{Assumption 1}, at the right hand side of (28), the term $\lambda_2\sum_{j=1}^k\frac{1}{m}(\sum_{1=1}^m\bigtriangleup_{ji})^2$ can be dropped.  Hence, 
\begin{equation}
    tr(\bigtriangleup^TL\bigtriangleup)\geq \lambda_2 ||\bigtriangleup||_F^2
\end{equation}

\section{Experimental validation }
\begin{figure}[htbp]
\floatconts
  {fig:fig_1}
  {\caption{Results of applying the Laplacian regularized estimator (3) and the ridge estimator (4)} to
  representation learning (2). (a) Plots both theoretical and empirical $||\hat{\Theta}-\Theta^*||_F$ versus the sample size $n$ for the Laplacian regularized estimator (3). (b) Plots both theoretical and empirical $||\hat{\Theta}-\Theta^*||_F$ versus the sample size $n$ for the ridge estimator (4).}
  {
    \subfigure[Laplacian regularized estimator][t]{
      \label{fig:fig_1a}
      \includegraphics[width=0.4\textwidth]{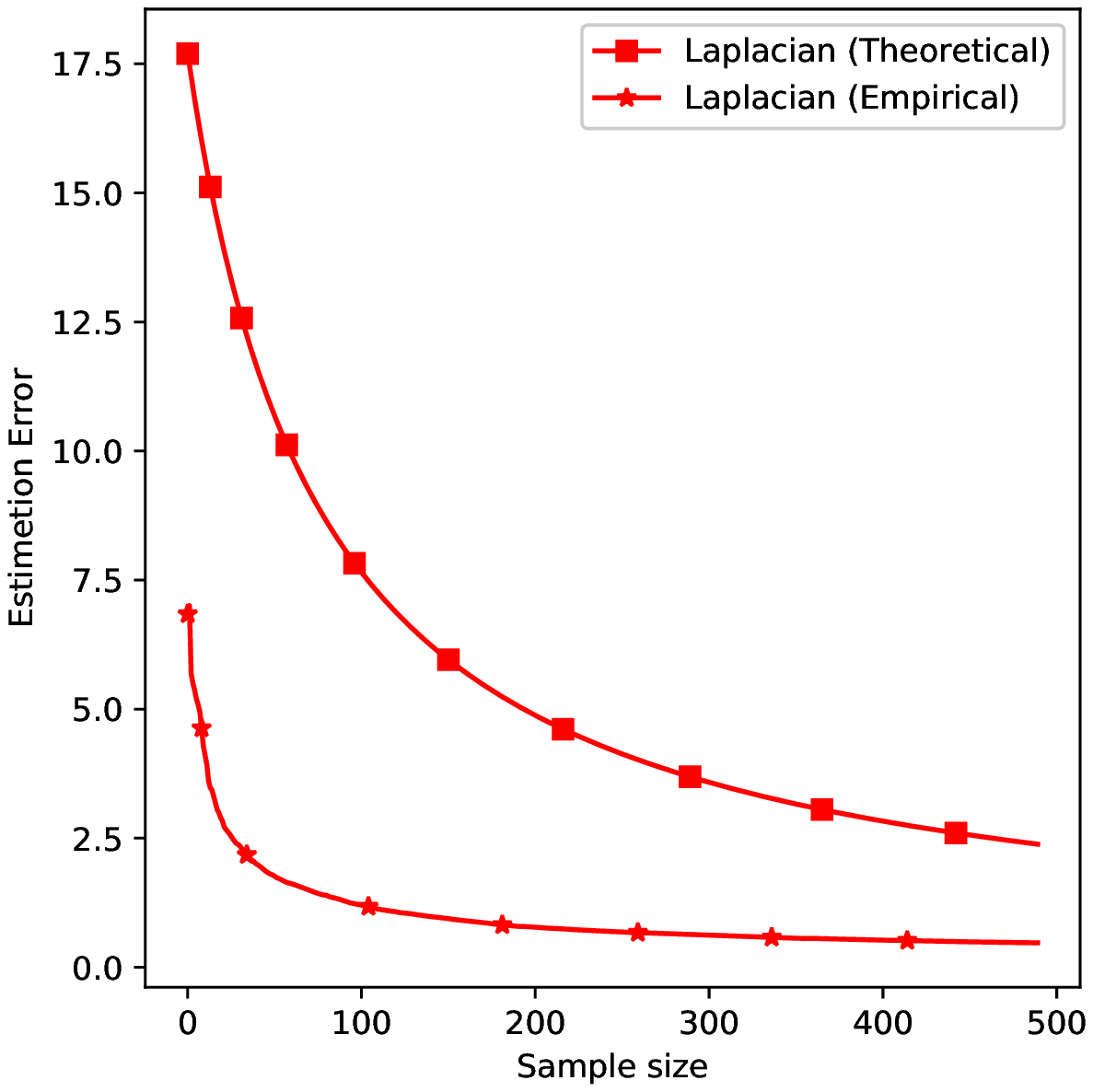}
    }\qquad 
    \subfigure[Ridge estimator][t]{
      \label{fig:fig_1b}
      \includegraphics[width=0.4\textwidth]{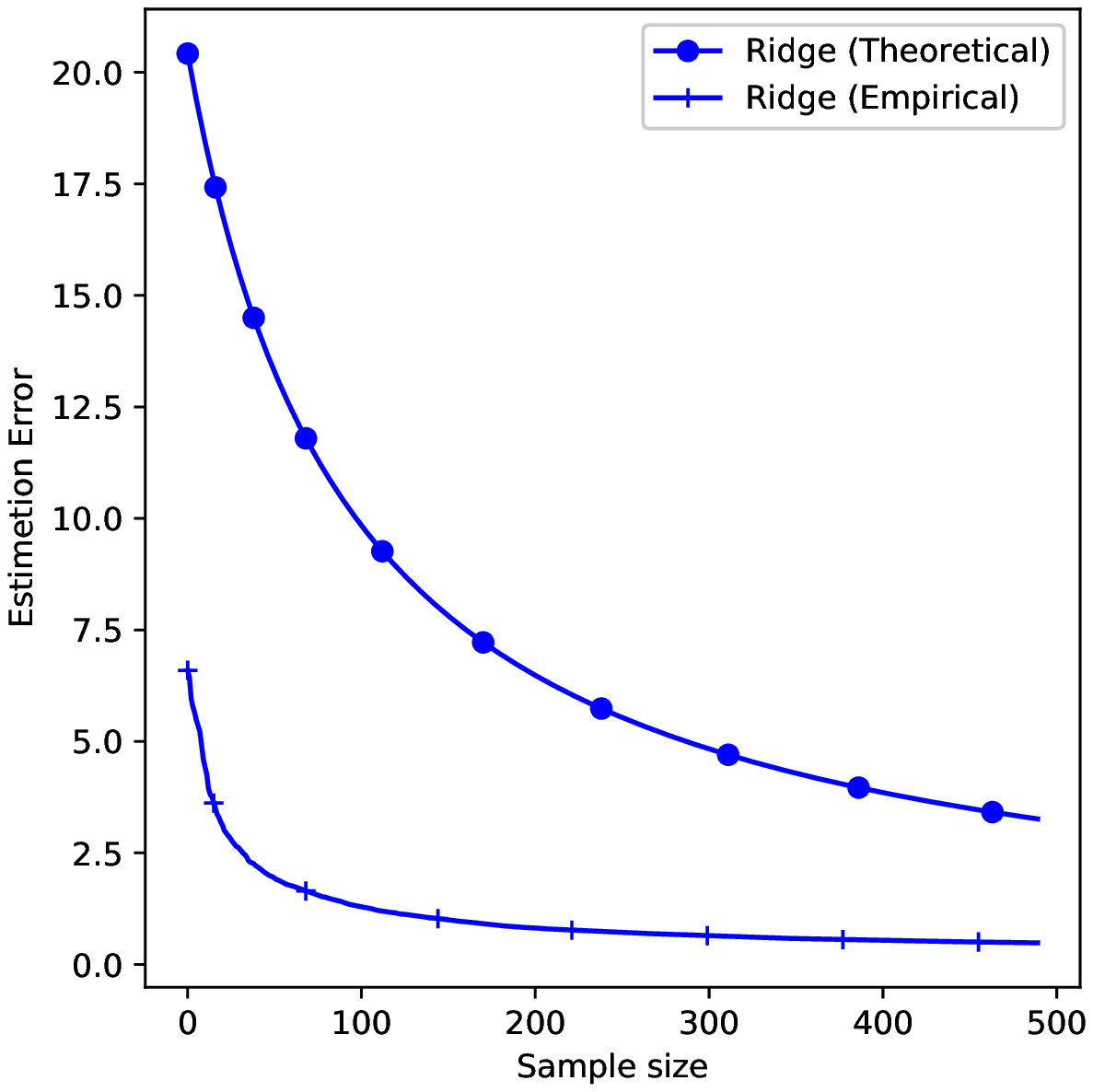}
} }
\end{figure}

We  present now simulation results that illustrate $i)$ the agreement between our theoretical results and empirical behaviors of the studied estimators provided in (3) and (4), $ii)$  a comparison between the two estimators to illustrate the impact of the graph structure on the estimation error.
For our simulations, we generate data according to the linear model (2), $Y=\Theta^*X+\Omega$ with $\Theta^*$ being smooth on $\mathcal{G}$, and then generate a random graph. Graph vertices' coordinates are generated uniformly at random in a unit square, then edge weights are computed by Gaussian radial basis function (RBF) (\textit{i.e.}, $exp(-d_{ij}^2/\sigma^2)$), where $d_{ij}$ is the distance between vertex $v_i$ and $v_j$, $\sigma=0.5$ is the kernel bandwidth parameter. With the Laplacian $L$, columns of $\Theta^*$ are generated by drawing samples from the multivariate Gaussian distribution $\mathcal{N}(0, L^{-1})$, where $L$ acts as the precision matrix. We then set the error $\Omega_{ij} \sim \mathcal{N}(0, 5)$. We set the number of sample $n=500$, the dimension of coefficient matrix $k=10$ and the size of $\Theta^*$, $m=100$.

Figure 1 shows the comparison between theoretical error bound from \textbf{Theorem 1} and empirical error $||\hat{\Theta}-\Theta^*||_F$ for the estimators provided in (3) and (4). Naturally, the error diminishes as the sample size $n$ grows, and more importantly the theoretical error bound depicts the empirical error bound versus the samples size $n$.
Finally, 
the Laplacian regularized estimator leads to better estimation by incorporating the graph structure. One fact should be mentioned that both estimators (3) and (4) are tuned to its best performance. The regularization parameter $\lambda$ of the graph Laplacian regularized estimator (3) is typically smaller than that of (4). This is expected since the regularizer $tr(\Theta^TL\Theta)$ is significantly larger than $tr(\Theta^TI_m\Theta)$.

\section{Conclusion}
In this paper, we have analyzed a graph Laplacian regularized estimator, and obtained a non-asymptotic error bound on the Frobenius norm of estimation error. We stated the bound as the main theorem, which provides a clear interpretation of the impact of both  graph structure and the smoothness prior  on the estimation accuracy. Finally, we provide empirical evidences that show good agreement with our theoretical analysis.
This theoretical analysis provides insights on the effectiveness of representation learning problems, naturally present in applications across many fields such as image denoising, compressed sensing, dictionary learning, and collaborative filtering (CF).


\newpage 
\bibliography{reference}

\newpage
\appendix

\section{Proof of (8) and (9)}
\subsection{Proof of (8)}
\begin{proof}
\begin{equation}
    \mathcal{L}(\Theta^*)=\frac{1}{2n}||Y-\Theta^* X||_F^2
\end{equation}
Thus, 
\begin{equation}
\begin{split}
    \bigtriangledown \mathcal{L}(\Theta^*)
    &=\frac{1}{n}(\Theta^*X-Y)X^T\\
    &=\frac{1}{n}\Omega X^T
\end{split}
\end{equation}
where $\Omega=\theta^*X-Y$.
\end{proof}
\subsection{Proof of (9)}
\begin{proof}
\begin{equation}
\begin{split}
    \delta \mathcal{L}(\Theta^*)
    &=\mathcal{L}(\Theta^*+ \bigtriangleup)-\mathcal{L}(\Theta^*)-\langle \bigtriangledown \mathcal{L}(\Theta^*),\bigtriangleup \rangle\\
    &=\frac{1}{2n}||Y-(\Theta^*+\bigtriangleup)X||_F^2-\frac{1}{2n}||Y-\Theta^*X||_F^2-\frac{1}{n}Tr(X\Omega^T\bigtriangleup)
\end{split}
\end{equation}
First, we expand the term $    ||Y-(\Theta^*+\bigtriangleup)X||^2_F-||Y-\Theta^*X||_F^2$ as
\begin{equation}
    Tr[(Y-(\Theta^*+\bigtriangleup )X)^T(Y-(\Theta^*+\bigtriangleup) X)]-Tr[(Y-\Theta^*X)^T(Y-\Theta^*X)]
\end{equation}
Which is 
\begin{equation}
    Tr[X^T\bigtriangleup^T \bigtriangleup X+2X^T(\Theta^*)^T\bigtriangleup X-2Y^T\bigtriangleup X]
\end{equation}
Next, we expand the last term in (32)
\begin{equation}
\begin{split}
    Tr(X\Omega^T\bigtriangleup)
    &=Tr(X(\Theta^*X-Y)^T\bigtriangleup)\\
    &=Tr(XX^T(\Theta^*)^T\bigtriangleup-XY^T\bigtriangleup)
\end{split}
\end{equation}
Substituting (34) and (35) into (32), we have $\delta \mathcal{L}(\Theta^*)$ be
\begin{equation}
    \frac{1}{2n}(Tr(X^T\bigtriangleup^T \bigtriangleup X))+\frac{1}{n}[Tr(X^T(\Theta^*)^T \bigtriangleup X)-Tr(Y^T\bigtriangleup X)+Tr(XY^T\bigtriangleup)-Tr(XX^T(\Theta^*)^T \bigtriangleup)]
\end{equation}
Due to the cyclic property of trace operator, terms of (36) are cancelled out except the first term. Therefore,
\begin{equation}
    \delta \mathcal{L}(\Theta^*)=\frac{1}{2n}Tr((\bigtriangleup X)^T(\bigtriangleup X))=\frac{1}{2n}||\bigtriangleup X||_F^2
\end{equation}
\end{proof}
\section{Proof of Theorem 1}
\begin{proof}
Due to the optimality of $\hat{\theta}$,
\begin{equation}
    \mathcal{L}(\hat{\Theta})+\alpha \mathcal{R}(\hat{\Theta}) \leq \mathcal{L}(\Theta^*)+\alpha \mathcal{R}(\Theta^*)
\end{equation}
Substituting $\hat{\theta}=\theta^*+\bigtriangleup$ and arrange the terms, we have
\begin{equation}
    \mathcal{L}(\Theta^*+\bigtriangleup)-\mathcal{L}(\Theta^*)+\alpha (\mathcal{R}(\Theta^*+\bigtriangleup)-\mathcal{R}(\Theta^*))\leq 0
\end{equation}
 Given $\mathcal{R}(\Theta)=Tr(\Theta^T L \Theta)$, We expand the term $\mathcal{R}(\Theta^*+\bigtriangleup)-\mathcal{R}(\Theta^*)$ as
\begin{equation}
\begin{split}
   \mathcal{R}(\Theta^*+\bigtriangleup)-\mathcal{R}(\Theta^*)
  &=Tr((\Theta^*+\bigtriangleup)^T L (\Theta^*+\bigtriangleup))-Tr((\Theta^*)^T L \Theta^*) \\
  &=Tr((\Theta^*)^T L \Theta^*)+(\Theta^*)^T L \bigtriangleup+\bigtriangleup^T L \Theta^*+\bigtriangleup^T L \bigtriangleup)-Tr((\Theta^*)^TL \Theta^*)\\
   &=2Tr((\Theta^*)^T L \bigtriangleup)+Tr(\bigtriangleup^T L \bigtriangleup)\\
   &=2Tr((\Theta^*)^T L\bigtriangleup)+\mathcal{R}(\bigtriangleup)\\
   &=2\langle L\Theta^*, \bigtriangleup \rangle+
   \mathcal{R}(\bigtriangleup)\\
   & \geq -2|\langle L \Theta^*, \bigtriangleup \rangle|+\mathcal{R}(\bigtriangleup)
\end{split}
\end{equation}
Substituting the last inequality into (39). we have
\begin{equation}
\mathcal{L}(\Theta^*+\bigtriangleup)-\mathcal{L}(\Theta^*)+\alpha(-2|\langle L \Theta^*, \bigtriangleup \rangle|+\mathcal{R}(\bigtriangleup))\leq 0
\end{equation}
From the definition of $\delta \mathcal{L}(\Theta^*)$ (7), we know that 
\begin{equation}
    \mathcal{L}(\Theta^*+\bigtriangleup)-\mathcal{L}(\Theta^*)=\langle \bigtriangledown \mathcal{L}(\Theta^*), \bigtriangleup \rangle +\delta \mathcal{L}(\Theta^*)
\end{equation}
Substituting this into (41), yields
\begin{equation}
    \langle \bigtriangledown \mathcal{L}(\Theta^*), \bigtriangleup \rangle +\delta \mathcal{L}(\Theta^*)+\alpha(-2|\langle L \Theta^*, \bigtriangleup \rangle|+\mathcal{R}(\bigtriangleup))\leq 0
\end{equation}
Note that
\begin{equation}
    \langle \bigtriangledown \mathcal{L}(\Theta^*), \bigtriangleup \rangle \geq -|\langle \bigtriangledown \mathcal{L}(\Theta^*), \bigtriangleup \rangle|
\end{equation}
With this, we have
\begin{equation}
    -|\langle \bigtriangledown \mathcal{L}(\Theta^*), \bigtriangleup \rangle|+\delta \mathcal{L}(\Theta^*)+\alpha(-2|\langle L \Theta^*, \bigtriangleup \rangle|+\mathcal{R}(\bigtriangleup))\leq 0
\end{equation}
Applying the H\"{o}lder Inequality \citep{kuptsov2001holder}, we have $    |\langle \bigtriangledown \mathcal{L}( \Theta^*), \bigtriangleup \rangle|\leq ||\bigtriangledown \mathcal{L}( \Theta^*)||_{\infty} ||\bigtriangleup||_*$ and $|\langle L \Theta^*, \bigtriangleup \rangle|\leq ||L\Theta^*||_F ||\bigtriangleup||_F$, thus
\begin{equation}
    \delta \mathcal{L}(\Theta^*, \bigtriangleup)+\alpha \mathcal{R}(\bigtriangleup)\leq ||\bigtriangledown\mathcal{L}(\bigtriangleup, \Theta^*)||_\infty||\bigtriangleup||_*+2\alpha ||L\Theta^*||_F ||\bigtriangleup||_F
\end{equation}
If we assume, $\alpha \geq ||\bigtriangledown \mathcal{L}(\bigtriangleup, \Theta^*)||_{\infty}$, we have
\begin{equation}
    \delta \mathcal{L}(\Theta^*, \bigtriangleup)+\alpha \mathcal{R}(\bigtriangleup)\leq \alpha||\bigtriangleup||_*+2\alpha ||L\Theta^*||_F ||\bigtriangleup||_F
\end{equation}
Due to \textbf{Lemma 1} $tr(\bigtriangleup^TL\bigtriangleup)\geq \lambda_2||\bigtriangleup||_F$ and the \textit{strong convexity} condition (10), $\delta \mathcal{L}(\Theta^*)\geq \kappa||\bigtriangleup||_F^2$, we have 
\begin{equation}
    \kappa||\bigtriangleup||_F^2+\alpha \lambda_2||\bigtriangleup||_F^2\leq \alpha ||\bigtriangleup||_*+2\alpha||L\Theta^*||||\bigtriangleup||_F
\end{equation}
Note the fact that if $rank(\bigtriangleup)\leq r$, then $||\bigtriangleup||_* \leq \sqrt{r}||\bigtriangleup||_F$. Substituting this into (48) and dividing both sides with $||\bigtriangleup||_F$ yields
\begin{equation}
    ||\bigtriangleup||_F \leq \frac{\alpha(\sqrt{r}+2||L\Theta^*||_F)}{\kappa+\alpha \lambda_2}
\end{equation}
\end{proof}
\section{Proof of Corollary 2}
The estimator is defined in (4) is a ridge estimator applied to estimate the design matrix $\Theta$. We denote it a ridge estimator below if not confusion introduced. 
\begin{equation}
    \hat{\Theta}_{ridge}=\arg \min_{\Theta \in \mathbb{R}^{m \times k}} \frac{1}{2n}||Y-\Theta X||_F^2+\alpha ||\Theta||_F^2
\end{equation}
Note that the ridge estimator is equivalent to 
\begin{equation}
    \hat{\Theta}_{ridge}=\arg \min_{\Theta \in \mathbb{R}^{m \times k}} \frac{1}{2n}||Y-\Theta X||_F^2+\alpha tr(\Theta^T I_m \Theta)
\end{equation}
Where $I_m$ is the identity matrix $I_m \in \mathbb{R}^{m \times m}$.  \\
By following the same arguments in Appendix B. We have 
\begin{equation}
    ||\hat{\Theta}_{ridge}-\Theta^*||_F \leq \frac{\alpha (\sqrt{r}+2||\Theta^*||_F)}{\kappa+\alpha}
\end{equation}
Note that $1=\lambda_1(I_m)=\lambda_2(I_m)=...=\lambda_m(I_m)$. So, there is no difference between employing $\lambda_1(I_m)$ or $\lambda_2(I_m)$.

\section{Justification of Assumption 1}
Given the definition $\bigtriangleup=\hat{\Theta}-\Theta^*$, it is reasonable to expect that entries of $\bigtriangleup_j$ varies around 0. \textit{i.e.}, the set $\{\bigtriangleup_{j1}, \bigtriangleup_{j2},...,\bigtriangleup_{jm}\}$ consists of both positive, negative real number and 0. Under such condition, it is reasonable to assume
$\sum_{i=1}^m\bigtriangleup_{ji}^2 \gg \frac{1}{m}(\sum_{i=1}^m\bigtriangleup_{ji})^2 $ since positive and negative real numbers would cancel out to some extent, which leads to a small $\frac{1}{m}(\sum_{i=1}^m \bigtriangleup_{ji})^2$, while $\sum_{i=1}^m\bigtriangleup_{ji}^2$ is not affected. \\
This pattern is expected to be consistent across columns of $\bigtriangleup$ so that we assume 
\begin{equation}
    \sum_{j=1}^k\sum_{i=1}^m\bigtriangleup_{ji}^2 \gg \sum_{j=1}^k\frac{1}{m}(\sum_{i=1}^m \bigtriangleup_{ji})^2
\end{equation}
\end{document}